\documentclass[10pt,twocolumn,letterpaper]{article}

\usepackage[pagenumbers]{cvpr} 

\definecolor{cvprblue}{rgb}{0.21,0.49,0.74}
\usepackage[pagebackref,breaklinks,colorlinks,allcolors=cvprblue]{hyperref}

\def\paperID{18782} 
\def\confName{CVPR}
\def\confYear{2025}

\title{PromptHash: Affinity-Prompted Collaborative Cross-Modal Learning for Adaptive Hashing Retrieval}

\author{Qiang~Zou\qquad Shuli~Cheng\thanks{Corresponding Author}\hspace{4pt}\qquad Jiayi~Chen \\
{School of Computer Science and Technology, Xinjiang University, China}\\
 zouq@stu.xju.edu.cn \qquad cslxju@xju.edu.cn \qquad 107552203982@stu.xju.edu.cn\\
}

\begin{document}
\maketitle
\begin{abstract}

\textit{Cross-modal hashing is a promising approach for efficient data retrieval and storage optimization. However, contemporary methods exhibit significant limitations in semantic preservation, contextual integrity, and information redundancy, which constrains retrieval efficacy. We present PromptHash, an innovative framework leveraging affinity prompt-aware collaborative learning for adaptive cross-modal hashing. We propose an end-to-end framework for affinity-prompted collaborative hashing, with the following fundamental technical contributions: (i) a text affinity prompt learning mechanism that preserves contextual information while maintaining parameter efficiency, (ii) an adaptive gated selection fusion architecture that synthesizes State Space Model with Transformer network for precise cross-modal feature integration, and (iii) a prompt affinity alignment strategy that bridges modal heterogeneity through hierarchical contrastive learning. To the best of our knowledge, this study presents the first investigation into affinity prompt awareness within collaborative cross-modal adaptive hash learning, establishing a paradigm for enhanced semantic consistency across modalities. Through comprehensive evaluation on three benchmark multi-label datasets, PromptHash demonstrates substantial performance improvements over existing approaches. Notably, on the NUS-WIDE dataset, our method achieves significant gains of $18.22\%$ and $18.65\%$ in image-to-text and text-to-image retrieval tasks, respectively. The code is publicly available at \url{https://github.com/ShiShuMo/PromptHash}.}
\end{abstract}
\section{Introduction}
\label{sec:intro}
With the rapid proliferation of social media platforms, multimodal data has grown exponentially, making cross-modal retrieval an increasingly promising field with broad application prospects~\cite{b1}. 

However, the inherent modal heterogeneity of multimodal data presents significant challenges, indicating substantial room for improvement in both academic research and industrial applications of cross-modal retrieval~\cite{b2}. Cross-modal hashing employs shared hash functions with strong representational capacities, mapping data from different modalities into a common hash space. As a result, data with similar semantic content across modalities can be mapped to proximate binary hash codes, effectively bridging the semantic gap between modalities and enabling rapid and accurate cross-modal data retrieval~\cite{b3,b12,b13,b14,b15,b17,b28,b29,b32}.
\begin{figure}[t]
\centering
\includegraphics[width=0.5\textwidth,height=5.5cm]{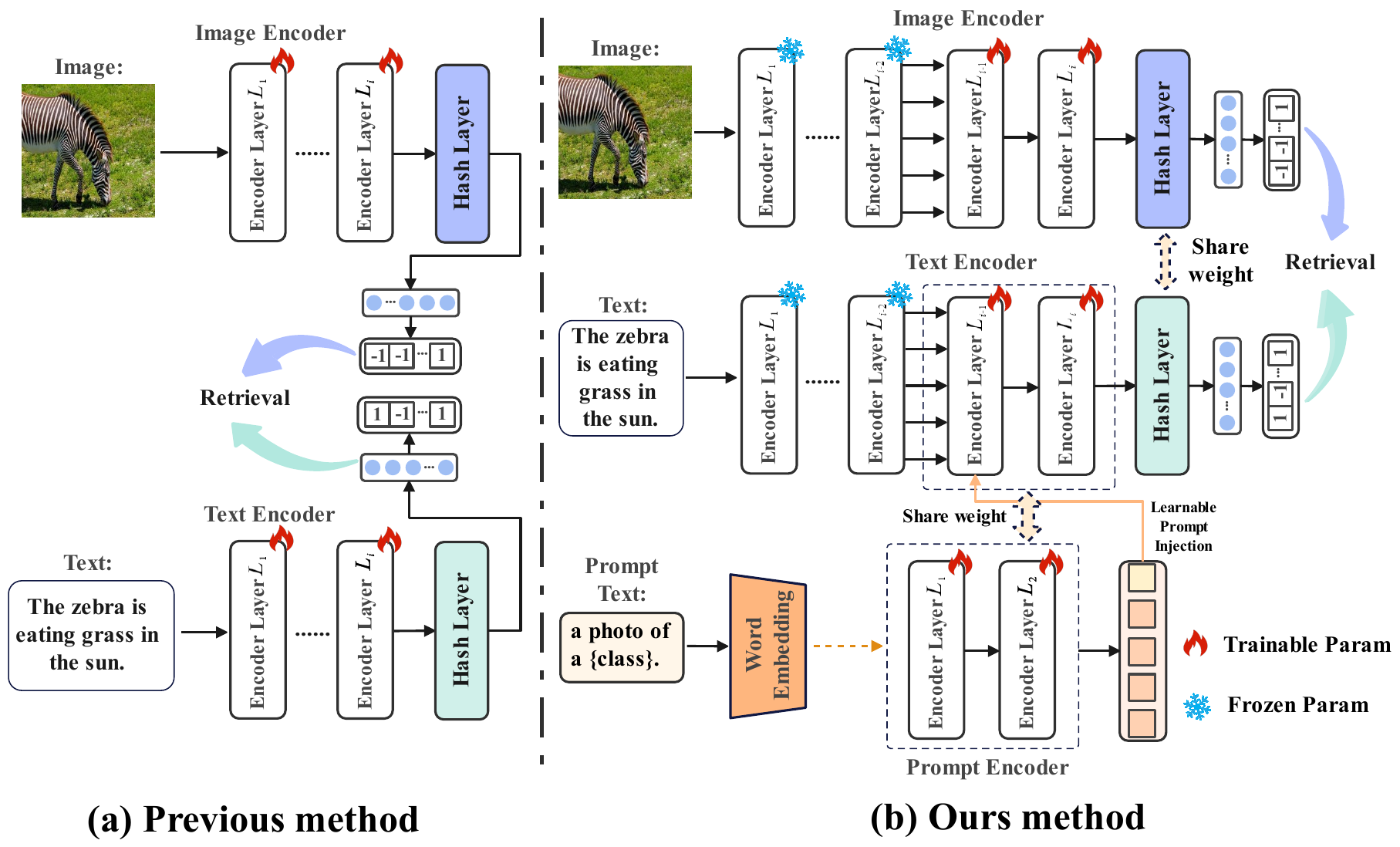}
\caption{Compare with existing frameworks. (a) Previous methods used dual Transformers for cross modal hashing contrastive learning. (b) We compare and learn images and texts separately by setting learnable affinity prompts.}
\label{figure1}
\end{figure}
Existing cross-modal hashing methods are generally divided into unsupervised and supervised approaches, with supervised methods typically outperforming unsupervised ones by leveraging supervisory information, such as pairwise similarity matrices or semantic labels. Despite the considerable progress made in deep supervised cross-modal hashing, there are still fundamental challenges to address.

Recently, CLIP (Contrastive Language-Image Pretraining)~\cite{b4} has emerged as a powerful backbone for cross-modal feature extraction in many hashing methods due to its strong zero-shot capabilities and alignment between visual and textual representations. However, adopting CLIP introduces several critical challenges that limit performance in real-world applications. 

A notable challenge is the context-length limitation in CLIP pre-trained models, where the current maximum context length is limited to 77 characters, whereas real-world text data often far exceeds this limit. This discrepancy leads to context loss and semantic truncation, especially when retrieving textual information. Furthermore, effectively fusing cross-modal features remains an open challenge, particularly in enhancing semantic feature weights relevant to retrieval while suppressing irrelevant information. For instance, the widely used MS COCO dataset~\cite{b5} exemplifies this challenge, as its text lengths range from 169 to 625 characters, significantly exceeding CLIP's 77-character capacity for text encoding. To address the CLIP context-length limitation, our approach incorporates foreground target text into adaptive affinity prompts and implements a dynamic weighting mechanism between these prompts and the original text. This strategy effectively preserves critical semantic information while circumventing CLIP's 77-character constraint, enabling comprehensive processing of lengthy real-world texts without semantic truncation. Previous approaches have employed various methods to address this: the Bag-of-Words (BoW) model, which disregards contextual word relationships; pre-trained Transformer~\cite{b6} models such as Vision Transformer(ViT)~\cite{b42} and BERT~\cite{b7}, which still face truncation; and VTPH's~\cite{b9} use of large language models~\cite{b8,b44} to reconstruct text as prompts, although this method introduces significant computational overhead without fully resolving the semantic truncation issue.

Another challenge lies in the feature fusion of different modalities, where existing cross-modal hashing methods often incorporate contrastive learning to capture inter-modal consistency. However, benchmark datasets commonly used in cross-modal retrieval suffer from context loss and semantic redundancy, especially in textual representation. For example, the textual content in MIRFLICKR-25K~\cite{b45} and NUS-WIDE~\cite{b46} is generated through direct label concatenation, resulting in limited contextual depth. Additionally, MS COCO's text data, derived from merged manual descriptions, can introduce semantic redundancy.

To address these challenges, we propose a Adaptive Affinity-Prompted Collaborative Cross-Modal Learning (PromptHash) for Hashing Retrieval in~\cref{figure1}. Ours framework that integrates text affinity prompts with State Space Model(SSM) and Transformer-based adaptive gated selection fusion, supported by a novel Prompt Affinity Contrastive Loss (PACL) to better align cross-modal semantics. Our contributions can be summarized as follows:

\begin{itemize}
\item To mitigate semantic truncation caused by CLIP's context-length limitation, adaptive text prompts are introduced that incorporate foreground target text essential for retrieval. This approach implements a dynamic weighting mechanism between prompts and original text, effectively preserving critical semantic information while circumventing length constraints for improved retrieval performance.

\item The framework also incorporates an SSM-Transformer adaptive gated selection fusion module to refine cross-modal feature integration, preserving relevant semantic information while filtering out redundant content. 

\item To balance prompt semantics with the original image-text semantics, a novel Prompt Affinity Contrastive Loss (PACL) is introduced, bridging modality heterogeneity and semantic gaps through global and local prompt contrast.

\item Extensive experiments on three benchmark datasets demonstrate that the proposed PromptHash framework significantly outperforms state-of-the-art methods, underscoring its effectiveness in cross-modal retrieval.
\end{itemize}

\section{Related Work}
\label{sec:related}
\subsection{Deep Cross-Modal Hashing}
Cross-modal hashing has made significant strides in information retrieval by mapping heterogeneous data into a unified Hamming space via hash functions. The literature typically categorizes approaches into unsupervised and supervised methods. 

Unsupervised methods maintain semantic similarity without explicit supervision by leveraging cross-modal correlations. Notable works include DAEH~\cite{b10} with deep adaptive enhancement, UKD's~\cite{b16} teacher-student framework, and UCMFH's~\cite{b18} CLIP-based feature extraction. Additionally, UCCH~\cite{b19} addresses binary-continuous relaxation constraints, while NRCH~\cite{b20} advances through robust contrastive loss and dynamic noise separation.

Supervised methods utilize label information to establish semantic consistency within a shared hash space. DCMH~\cite{b3} pioneered end-to-end deep cross-modal hashing, followed by CMHH's~\cite{b11} focal loss approach and AGAH's~\cite{b12} adversarial learning with attention mechanisms. Recent innovations include DCHUC's~\cite{b21} joint optimization, MIAN's~\cite{b22} modality-invariant networks, and GCDH's~\cite{b23} graph convolutional networks. DCHMT~\cite{b24} and MITH~\cite{b25} leverage CLIP for fine-grained feature extraction.

Proxy learning has emerged as another significant direction, with DAPH's~\cite{b26} data-aware networks, DSPH's~\cite{b27} fine-grained semantic relations, and DHaPH's~\cite{b30} hierarchical learning. Notable approaches also include CMCL's~\cite{b31} multi-bit collaboration and VTPH's~\cite{b9} large model optimization. Distinct from existing methods, we integrate both global and local prompt alignments while minimizing feature divergence to alleviate modality heterogeneity and semantic gaps.

\subsection{Prompt Learning}
Prompt learning originated in NLP, integrating handcrafted templates into input data for downstream tasks. Recent studies have extended this concept to vision-language models, with CoCoOp~\cite{b33} introducing Conditional Context Optimization, MaPLe~\cite{b34} implementing Multi-modal Prompt Learning, and CoPrompt~\cite{b35} enhancing performance through consistency constraints. While PromptKD~\cite{b36} explores Unsupervised Prompt Distillation, prompt learning's effectiveness in cross-modal hashing remains underexplored.

\subsection{Contrastive Learning}
Contrastive learning has demonstrated substantial efficacy in feature representation by maximizing similarity between positive pairs while minimizing negative pair similarity. Several cross-modal hashing methods have adopted this paradigm: UCCH~\cite{b19} pioneered unsupervised contrastive learning, UniHash~\cite{b37} facilitates instance-level learning, and CMCL~\cite{b31} optimizes through token alignment. However, existing methods often overlook explicit fine-grained token-level semantic alignment. Our work addresses this through prompt-based contrastive learning for both instance and token-level representations.

\subsection{State Space Models}
While contrastive learning effectively aligns semantic representations across modalities, distinguishing between foreground (relevant) and background (irrelevant) information in multimodal contexts remains challenging. State Space Models (SSM)~\cite{b38} offer a promising solution through their selective sequence modeling capabilities with linear computational complexity. Unlike Transformers that attend to all tokens equally, SSM's selective scan mechanism can dynamically focus on informative elements while filtering out noise. Vim~\cite{b39} pioneered SSM in visual domains, demonstrating their ability to selectively process visual information. VMamba~\cite{b40} extended this capability with 2D Selective Scanning that effectively separates foreground objects from background elements. Jamba~\cite{b41} further showed how SSM's selective attention, when integrated with Transformer layers, enables more efficient sequence feature processing. 

Despite these advances, the potential of SSM, when combined with contrastive learning, to effectively address the foreground-background confusion problem in cross-modal hashing retrieval remains unexplored. Our work integrates these two paradigms, utilizing SSM for precise separation of foreground and background information and contrastive mechanisms for accurate cross-modal alignment.

\section{Methodology}
\label{sec:methodology}
\subsection{Notations and Problem Definition}
This paper adopts a framework analogous to conventional cross-modal hashing methodologies, utilizing image-text pairs as input modalities. Let $\mathcal{O} = \{o_i\}_{i=1}^N$ denote a collection of $N$ training sample pairs, where each sample $o_i = (x_i^v, x_i^t, l_i, p_i)$ comprises a visual feature vector $x_i^v \in \mathbb{R}^{d^v}$, a textual feature vector $x_i^t \in \mathbb{R}^{d^t}$, a multi-label annotation $l_i \in \{0,1\}^{1 \times C}$ (where $l_{i,c} = 1$ if sample $o_i$ belongs to category $c$; otherwise, $l_{i,c} = 0$), and a prompt word $p_i$. We construct a similarity matrix $S$ where $S_{ij} = 1$ if samples share at least one category, indicating semantic affinity; otherwise, $S_{ij} = 0$. The primary objective is to project heterogeneous features into a unified $K$-bit Hamming space through two hashing functions: $b_i^v = H^v(f_i^v; \theta^v) \in \{-1, +1\}^K$ and $b_i^{pt} = H^t(f_i^{pt}; \theta^t) \in \{-1, +1\}^K$, where $b_i^v$ and $b_i^{pt}$ denote the hash codes for visual and textual modalities, while $\theta^v$ and $\theta^t$ represent the learnable parameters of the corresponding functions. Throughout this paper, $M$ denotes the mini-batch size, and $D$ represents the feature embedding dimension. We use $f$ for initial feature representations (e.g., $f_i^v$, $f_i^t$, $f_i^p$, $f_i^{pt}$), $h$ for continuous hash representations prior to quantization (e.g., $h_i^v$, $h_i^{pt}$, $h_i^p$), and $b$ for binary hash codes(e.g., $b_i^v$, $b_i^{pt}$). For similarity metrics, $\Theta_{ij}$ represents prompt-visual similarity, $\Phi_{ij}$ denotes visual-prompt similarity, and $\Omega_{ij}$ indicates intra-modal prompt similarity.

\subsection{Feature Extraction}
\begin{figure*}
\centerline{\includegraphics[width=\textwidth]{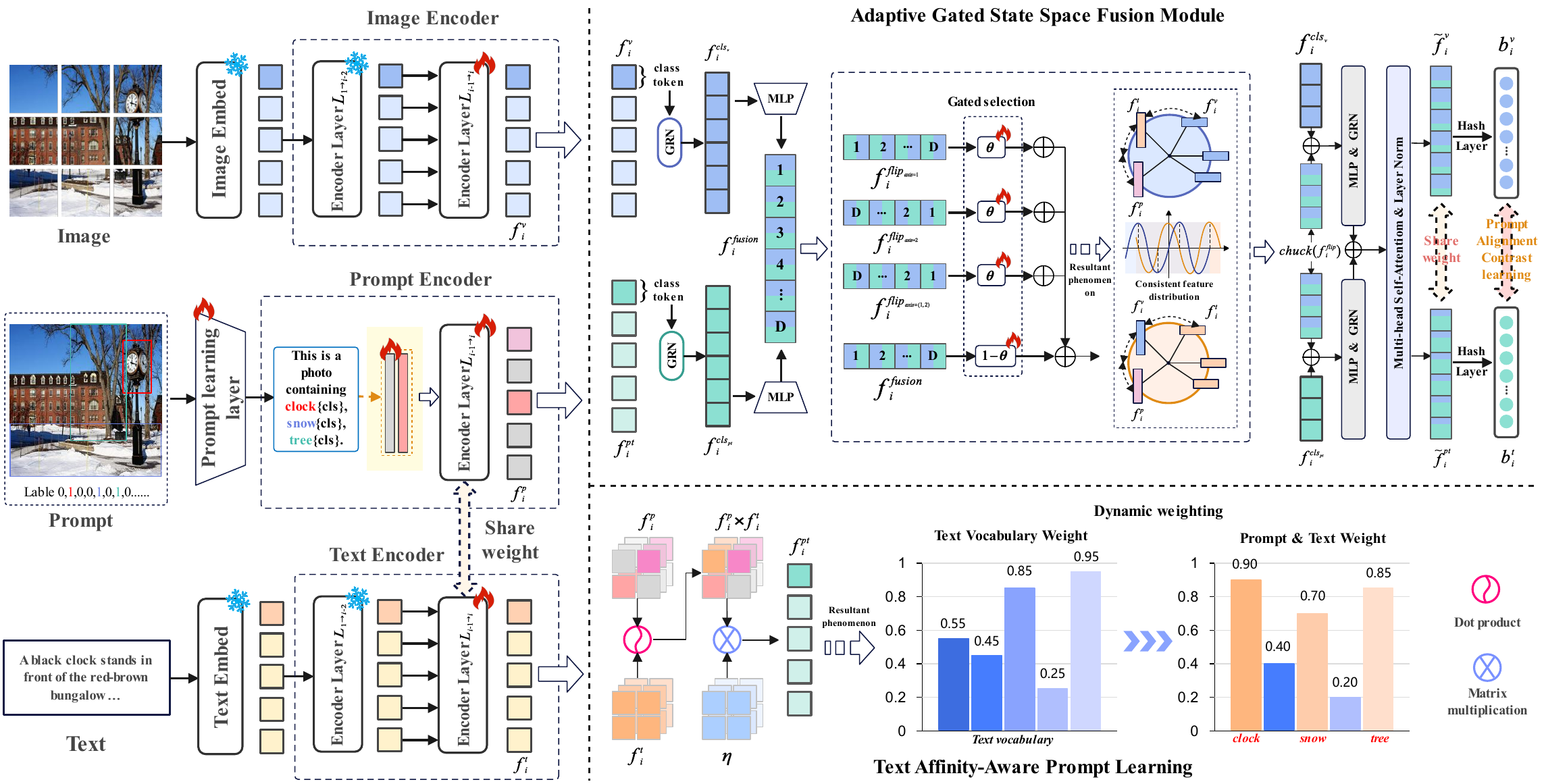}}
\caption{Overall Framework of PromptHash. Our framework consists of five key components: 1) Image and Text Encoders for modality-specific feature extraction; 2) Adaptive Gated State Selection and Fusion Module for feature filtering and cross-modal fusion between image features and hybrid prompt-enhanced textual features; 3) Text Affinity-Aware Prompting that dynamically learns and distinguishes retrieval-beneficial foreground information while optimizing textual feature representations through dynamic prompting mechanisms; 4) Cross-Modal Prompt Alignment Mechanism incorporating both global and local alignments, where global alignment facilitates image-to-text and image-to-prompt representation alignments with intra-class and inter-class affinity losses; 5) Hash Learning with quantization and reconstruction losses.}
\par\vspace{-2mm}
\label{figure2}
\end{figure*}

\cref{figure2} illustrates the detailed workflow of our proposed PromptHash framework. To leverage the latest advances in deep neural networks and Transformer-based models, we employ a two-branch Transformer architecture for the extraction of image and text features. Additionally, we introduce a dedicated Transformer network designed to extract prompt features, which are then used to compute prompt-weighted similarities. To further enhance the quality of the image and text representations, we utilize pretrained ViT and BERT as the visual encoder and text encoder, respectively. These models extract the image and text embeddings, denoted as $f_i^v = \{G_i^v\} \in \mathbb{R}^{L^\nu \times D}$ and $f_i^t = \{G_i^t\} \in \mathbb{R}^{L^t \times D}$. Similarly, the constructed masked fused prompt embeddings, denoted as $f_i^{pt} = \{G_i^p\} \in \mathbb{R}^{L^p \times D}$, are used to represent the fused prompt.
\subsection{Text Affinity-Aware Prompt Learning}
In cross-modal hashing tasks, negative textual features can significantly degrade retrieval performance by hindering the effective distinction between foreground and background information. To address this issue, we propose a Text Affinity-Aware Prompt (TAAP) module, which dynamically learns and differentiates foreground information that aids retrieval, and optimizes textual feature representations through a dynamic prompt mechanism. 

Let $l_i$ represent the category label set corresponding to image-text pairs $o_i$ and $o_t$. We generate prompt text using a template: ``This is an image containing $\{\text{class\_name}_i\}$," where $\{\text{class\_name}_i\}$ corresponds to label $l_i$. The complete prompt text is formulated as $\text{prompt}_i = \text{``This is an image containing \{class\_name\_i\}"}$, which is then tokenized using Byte Pair Encoding (BPE): $\text{token\_prompt}_i = \text{Tokenizer}(\text{prompt}_i)$.

The token sequence is encoded into an embedding vector $p_i$:
\vspace{-4mm}
\small
\begin{gather}
\begin{aligned}
\scalebox{0.9}{$
p_i = \text{embed}(\text{token\_prompt}_i), \quad p_i \in \mathbb{R}^{D}$
}
\end{aligned}
\label{eq1}
\end{gather}
where $D$ denotes CLIP's embedding dimension. Then ${f_i^p}$ combine this with a learnable context vector $C_{\text{ctx}}$:
\vspace{-1mm}
\small
\begin{gather}
\begin{aligned}
\scalebox{0.9}{$
{f_i^p} = p_i + C_{\text{ctx}}$}
\end{aligned}
\label{eq2}
\end{gather}
This dynamic prompt mechanism enables effective distinction between foreground and background features through a learnable mask, optimizing the feature representation for retrieval tasks. The generated prompt features $f_i^p$ are then processed by the TAAP module's Transformer encoder layers, fusing them with the original text features $f_i^t$:
\vspace{-2mm}
\small
\begin{gather}
\begin{aligned}
\scalebox{0.95}{$
\widetilde{f_i^{p}} = \text{TransformerLayer}({f_i^p})$}
\end{aligned}
\label{eq3}
\end{gather}

Finally, we obtain the weighted feature representation through element-wise multiplication and global average pooling:
\vspace{-1mm}
\small
\begin{gather}
\begin{aligned}
f_i^{pt} = (\widetilde{f_i^p} \times f_i^t)\otimes\eta, \quad f_i^{pt} \in \mathbb{R}^{D}
\end{aligned}
\label{eq4}
\end{gather}
where $\eta$ represent learnable weight, respectively, with $\otimes$ denoting matrix multiplication. This integration process effectively preserves foreground information while suppressing background noise, significantly improving cross-modal hashing retrieval performance.

\subsection{Adaptive Gated State Space Fusion Module}
In this work, we integrate the State Space Model (SSM) with adaptive gating mechanisms to design an adaptive filtering feature fusion module, which we refer to as the Adaptive Gated State Space Fusion Module(AGSF). The textual features are weighted through the aforementioned prompt learning module, yielding a refined textual feature representation. However, image features must also undergo filtering and fusion with the resulting mixed prompt-text features. This process is formalized as follows:
\vspace{-1mm}
\small
\begin{gather}
\begin{aligned}
f_i^{\text{cls}_m} &= \text{SiLU}(\text{MLP}(\text{GRN}(f_i^m))) \mid m \in\{v, pt\} \\
f_i^{\text{fusion}} &= \text{Concat}(f_i^{\text{cls}_v}, f_i^{\text{cls}_{pt}})
\end{aligned}
\label{eq5}
\end{gather}
where $\text{Concat}$ represents the concatenation operation, $\text{GRN}$ denotes Global Response Normalization, $\text{MLP}$ refers to a multi-layer perceptron, and $\text{SiLU}$ is the activation function. In the \textit{Global Response Normalization (GRN)} layer, for input $x \in \mathbb{R}^{B \times N \times D}$, the L2 norm $G_x$ and normalization factor $N_x$ are computed sequentially:
\vspace{-1mm}
\small
\begin{gather}
\begin{aligned}
G_x &= \| x \|_2 = \sqrt{\sum_{d=1}^{D} x_d^2}, \quad N_x = \frac{G_x}{\frac{1}{N} \sum_{i=1}^{N} G_{x_i}}
\end{aligned}
\label{eq6}
\end{gather}
The final output with learnable parameters $\lambda$ and $\kappa$ is:
\vspace{-1mm}
\begin{gather}
\scalebox{0.9}{$
\text{GRN}(x) = \lambda \cdot (x \cdot N_x) + \kappa + x
\label{eq7}$}
\end{gather}
After obtaining the fused features, they are fed into the SSM adaptive selection module, expressed as:
\vspace{-1mm}
\small
\begin{gather}
\begin{aligned}
f_i^{\text{flip}_{\text{axis}=\psi}} &= \text{Flip}_{\text{axis}=\psi}^T(\text{SSM}(\text{Flip}_{\text{axis}=\psi}(f_i^{\text{fusion}})))
\end{aligned}
\label{eq8}
\end{gather}
where $\psi \in \{1, 2, (1,2)\}$. Here, $\text{Flip}_{\text{axis}=\psi}$ denotes a flipping operation along different axes, and $\text{Flip}^T$ represents its inverse. The term SSM refers to the State Space Model. After obtaining the final three features, we apply an adaptive weighted filtering process. The parameter $\tau$ serves as a learnable temperature coefficient, which balances the relative importance of the four features derived from flipped and unflipped orientations:
\vspace{-1mm}
\small
\begin{gather}
\begin{aligned}
f_i^{\text{fit}} = \theta\tau\sum_{j \in \omega} f_i^{\text{flip}_{\text{axis}=j}} + (1-\theta)f_i^{\text{fusion}},\ \omega = \{1, 2, (1,2)\}
\end{aligned}
\label{eq9}
\end{gather}
Where, $\theta$ denotes a set of learnable parameters. The filtered features are subsequently passed through a Transformer encoder layer, yielding:
\vspace{-1mm}
\small
\begin{gather}
\begin{aligned}
\widetilde{f_{i}^v}, \widetilde{f_{i}^{\text{pt}}} = \text{Split}(\text{TransformerLayer}(\text{MLP}(f_{i}^{\text{fit}})))
\end{aligned}
\label{eq10}
\end{gather}
The resulting features $\widetilde{f_{i}^v}$ and $\widetilde{f_{i}^{\text{pt}}}$ represent the outputs obtained through the adaptive SSM gating and selection fusion module. Here, $\text{Split}$ denotes an operation that separates the individual visual and textual features from the fused representation. Through the aforementioned operations, we simultaneously adaptively filter and dynamically weight redundant information in both images and text, significantly enhancing the quality and effectiveness of retrieval.

\subsection{Prompt Alignment Contrastive Learning}
To effectively bridge the semantic gaps between different modalities, we propose prompt alignment contrastive learning (PACL), a learning mechanism that enables fine-grained alignment between modalities. This mechanism operates through two key components: cross-modal prompt alignment and affinity-aware loss function.

\subsubsection{Cross-Modal Prompt Alignment}
To further mitigate heterogeneity and semantic gaps, we leverage contrastive learning to explicitly align the fine-grained semantic concept representations. Our alignment strategy comprises two key components: global-local prompt alignment and affinity-aware loss functions.

For global prompt alignment, we adopt symmetric InfoNCE losses to align image representations with textual and prompt representations:
\vspace{-1mm}
\small
\begin{gather}
\begin{aligned}
\mathcal{L}_{i}^{a \to b} &= -\log \frac{\exp \left(\left(\widetilde{{f}_{i}^{a}}\right)^{T} \widetilde{{f}_{i}^{b}} / \tau_{1}\right)}{\sum_{c=1}^{M} \exp \left(\left(\widetilde{{f}_{i}^{a}}\right)^{T} \widetilde{{f}_{c}^{b}} / \tau_{1}\right)} \\
\mathcal{L}_{\text{gpa}} &= \frac{1}{M} \sum_{i=1}^{M} \sum_{\substack{a, b \in \{v, pt\} \\ a \neq b}} \mathcal{L}_i^{a \to b}
\end{aligned}
\label{eq11}
\end{gather}

where $\tau_{1}$ is a temperature hyperparameter, and the subscript $c$ indexes over all samples in the batch. Here, $a,b \in \{v,pt\}, a \neq b$, corresponding to visual and textual features.

For local prompt alignment, we employ a dynamic temperature mechanism adjusted by modality distributions:
\vspace{-1mm}
\small
\begin{gather}
\begin{aligned}
\tau_{2} = \tau \times \frac{1}{1 + \text{JS}\text{-}\text{div}({f}_{i}^{v},{f}_{i}^{p})}
\end{aligned}
\label{eq12}
\end{gather}
where $\tau$ is the base temperature and JS-div measures the Jensen-Shannon divergence between modality distributions. Here, $\tau$ and $\tau_1$ denote the temperature hyperparameters with a default value of 0.07.

The local alignment losses are then formulated as:
\vspace{-1mm}
\small
\begin{gather}
\begin{aligned}
\mathcal{L}_{i}^{e \to f} &= -\log \frac{\exp \left(\left(\widetilde{{f}_{i}^{e}}\right)^{T} \widetilde{{f}_{i}^{f}} / \tau_{2}\right)}{\sum_{c=1}^{M} \exp \left(\left(\widetilde{{f}_{i}^{e}}\right)^{T} \widetilde{{f}_{c}^{f}} / \tau_{2}\right)} \\
\mathcal{L}_{\text{lpa}} &= \frac{1}{M} \sum_{i=1}^{M} \sum_{\substack{e, f \in \{v, p\} \\ e \neq f}} \mathcal{L}_i^{e \to f}
\end{aligned}
\label{eq13}
\end{gather}
where $e,f \in \{v,p\}, e \neq f$, corresponding to visual and prompt features. $\tau$ and $\tau_1$ denote the temperature hyperparameters with a default value of 0.07.

\subsubsection{Affinity-Aware Loss Function}
To preserve semantic consistency, we design inter-class and intra-class affinity losses:
\vspace{-1mm}
\small
\begin{gather}
\begin{aligned}
\mathcal{L}_{\text{inter}} = &-\frac{1}{MN}\sum_{i=1}^{N} \sum_{j=1}^{M}\sum_{k=1}^{Q}\left(\mathcal{S}_{ij} k-\log \left(1+e^{\Theta_{ij}}\right)\right)
\end{aligned}
\label{eq14}
\end{gather}
where $Q \in \{\Theta_{ij}, \Phi_{ij}\}$, $\Theta_{ij} \neq \Phi_{ij}$, $\Theta_{ij} = \frac{1}{2} \left( h_{i}^{p} \right)^{T} h_{j}^{v}$ and $\Phi_{ij} = \frac{1}{2} \left( h_{i}^{v} \right)^{T} h_{j}^{p}$ denote cross-modal similarities, where $h_{i}^{p}$ and $h_{j}^{v}$ represent the real-valued hash codes of prompt and visual features respectively, before passing through the sign function, $N$ is the number of samples, and $\mathcal{S}_{ij} \in \{0,1\}$ is the ground truth similarity matrix.

The intra-class affinity loss is defined as:
\vspace{-1mm}
\small
\begin{gather}
\begin{aligned}
\mathcal{L}_{\text{intra}} = -\frac{1}{MN} \sum_{i=1}^{N} \sum_{j=1}^{M} \left( \mathcal{S}_{ij} \Omega_{ij} - \log \left(1 + e^{\Omega_{ij}}\right) \right)
\end{aligned}
\label{eq15}
\end{gather}
where $\Omega_{ij} = \frac{1}{2} \left( h_{i}^{p} \right)^{T} h_{j}^{p}$ represents intra-modal similarity between prompt features.

\subsection{Hashing Learning}
\textbf{1) Quantization Loss:} The quantization loss aims to learn a unified semantic representation and generate high-quality, distinguishable hash codes. It is mathematically formulated as follows:
\vspace{-1mm}
\small
\begin{gather}
\begin{aligned}
\scalebox{0.9}{$
\mathcal{L}_{\text{quan}} = \frac{1}{NM} \sum_{i=1}^{M} \left( \left\| b_{i}^v - \frac{1}{2} \left( h_{i}^{v} + f_{i}^{v} \right) \right\|_2^2 + \left\| b_{i}^{pt} - \frac{1}{2} \left( h_{i}^{pt} + f_{i}^{pt} \right) \right\|_2^2 \right)$}
\end{aligned}
\label{eq16}
\end{gather}
This loss function encourages the hash codes $b_i^v$ and $b_i^{pt}$ to closely approximate the average of the visual and textual feature representations, facilitating a unified and discriminative semantic embedding.

\textbf{2) Reconstruction Loss:} The reconstruction loss seeks to enhance the representational capacity and retrieval performance of the hash codes by ensuring that the generated hash codes are more distinguishable and closely approximate the original semantic features. The loss is expressed as:
\vspace{-1mm}
\small
\begin{gather}
\begin{aligned}
\mathcal{L}_{\text{recon}} = \frac{1}{M} \sum_{i=1}^{M} \left( \left\| h_i^{v} - b_i^{v} \right\|_2^2 + \left\| h_i^{pt} - b_i^{pt} \right\|_2^2 \right)
\label{eq17}
\end{aligned}
\end{gather}
The reconstruction loss minimizes the difference between the hash codes and their corresponding feature representations, promoting the generation of hash codes that effectively retain semantic information for accurate retrieval.

Finally, the total loss function for the proposed PromptHash method is computed as the weighted sum of the individual loss components, $\alpha, \beta, \gamma, \mu, \sigma, \zeta$ represent hyperparameters:
\vspace{-1mm}
\small
\begin{gather}
\begin{aligned}
\mathcal{L}_{\text{Total}} &= \alpha \mathcal{L}_{\text{gpa}} + \beta \mathcal{L}_{\text{lpa}} + \gamma \mathcal{L}_{\text{inter}} \\
&+ \mu \mathcal{L}_{\text{intra}} + \sigma \mathcal{L}_{\text{quan}} + \zeta \mathcal{L}_{\text{recon}}
\end{aligned}
\label{eq18}
\end{gather}

\section{Experiments}
\label{sec:exp}
To rigorously evaluate the effectiveness of the proposed PromptHash method, we conducted extensive experiments on three widely used cross-modal multi-label retrieval datasets: MIRFLICKR-25K, NUS-WIDE, and MS COCO. We compared PromptHash against twelve state-of-the-art cross-modal hashing retrieval methods, including DCMH~\cite{b3}, CMHH~\cite{b11}, GCDH~\cite{b23}, DCHMT~\cite{b24}, MITH~\cite{b25}, DSPH~\cite{b27}, TwDH~\cite{b28}, DNpH~\cite{b29}, DHaPH~\cite{b30}, CMCL~\cite{b31}, and VTPH~\cite{b9}. All methods were tested under identical experimental conditions, with consistent dataset splits, retrieval, and query sets aligned with our experimental setup.

The following sections provide a comprehensive analysis of the experimental results obtained from each of the eleven competing algorithms. Additionally, we offer detailed descriptions of the three datasets used for training and testing, outline the experimental procedures specific to PromptHash, and discuss the metrics used to assess its performance. The experimental environment specifications are also documented to ensure reproducibility and transparency.

\subsection{Datasets}
\cref{table1} presents an overview of the datasets used in our experiments and details their respective partitioning strategies. For our evaluations, we applied a consistent sampling strategy across the three large-scale, multi-label datasets: MIRFLICKR-25K, NUS-WIDE, and MS COCO. Each dataset was partitioned into training, testing, and retrieval sets. Image and text data were processed uniformly across all datasets: images were resized to an input resolution of 224 × 224 pixels, while text data were encoded using byte pair encoding (BPE) before being input into the network.

\subsection{Experimental Details}
In this study, we employ CLIP-B16 as the backbone network, with experiments conducted on a single NVIDIA RTX 4090 GPU (24GB) using PyTorch V2.3.1. Input images across all datasets are resized to ${224 \times 224}$ pixels. The learning rate is set to ${1\mathrm{e}{-6}}$ for the backbone network and ${1\mathrm{e}{-5}}$ for both the prompt model and fusion module, with a batch size of 128. The loss terms are defined as follows: ${\alpha}$ represents the global prompt alignment loss, ${\beta}$ denotes the local prompt alignment loss, ${\gamma}$ indicates the inter-class affinity loss, ${\mu}$ represents the intra-class affinity loss, ${\sigma}$ denotes the quantization loss, and ${\zeta}$ represents the reconstruction loss. For the MIRFLICKR-25K and NUS-WIDE datasets, these hyperparameters are set to ${5.0}$, ${5.0}$, ${0.005}$, ${5.0}$, ${0.1}$, and 0.001, respectively. For the MS COCO dataset, the corresponding values are ${5.0}$, ${5.0}$, ${0.005}$, ${20.0}$, ${1.0}$, and 0.001. Throughout all tables, bold typeface indicates the best performance, while underlined values represent the second-best results.

We conduct comprehensive ablation studies to systematically evaluate the effectiveness of each component in our proposed PromptHash framework, with results presented in~\cref{table2}. The experiments examine five distinct variants: (a) a baseline implementation utilizing only the CLIP feature extraction network and hash function, excluding TAAP, AGSF, and PACL modules; (b) PromptHash w/o (PACL + AGSF), which retains only the TAAP module; (c) PromptHash w/o (TAAP + PACL), maintaining only the AGSF module; (d) PromptHash w/o AGSF, preserving TAAP and PACL modules; and (e) PromptHash w/o PACL, retaining TAAP and AGSF modules.
\begin{figure*}
    \centering
    \includegraphics[width=\textwidth]{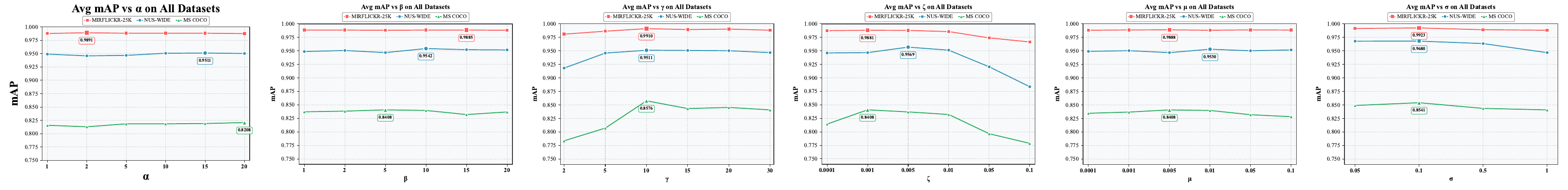}
    \caption{\label{figure3}Ablation study results of six key hyperparameters ($\alpha$, $\beta$, $\gamma$, $\lambda$, $\mu$, $\nu$) evaluated on three benchmark datasets (MIRFLICKR-25K, NUS-WIDE, and MS COCO).}
    \par\vspace{-3mm}
\end{figure*}
\begin{figure*}
    \centering
    \includegraphics[width=\textwidth]{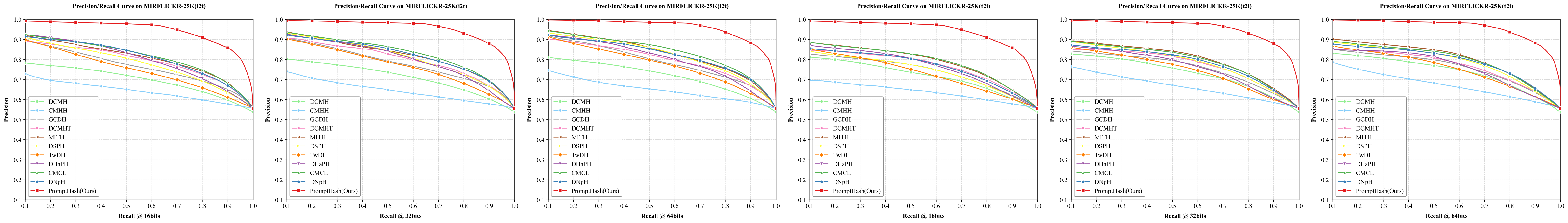}
    \includegraphics[width=\textwidth]{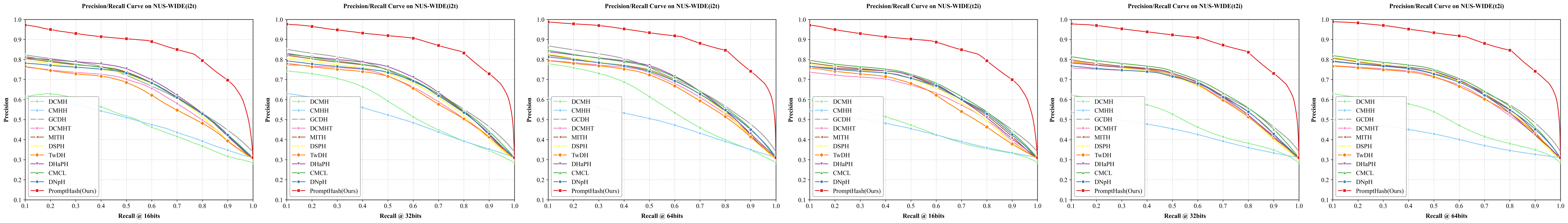}
    \includegraphics[width=\textwidth]{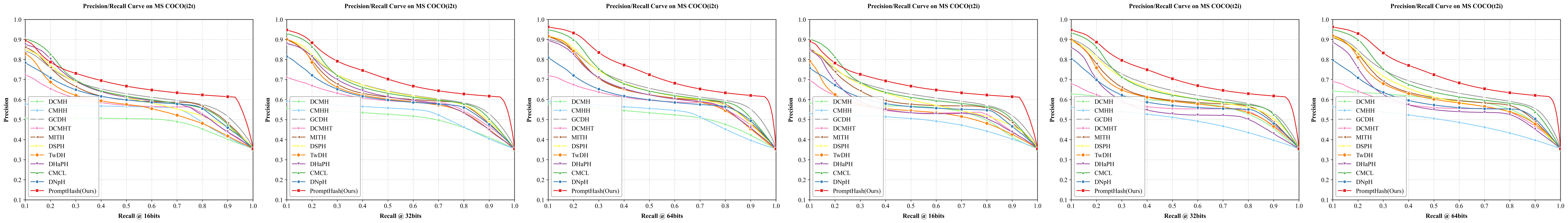}
    \caption{\label{figure4}Precision-Recall curves of different hash code lengths (16, 32, and 64 bits) on three benchmark datasets: MIRFLICKR-25K, NUS-WIDE, and MS COCO.}
    \par\vspace{-3mm}
\end{figure*}
\begin{table}[!htbp]
\centering
\caption{\label{table1}Summary of Feature Statistics for the Three Benchmark Datasets}
\resizebox{0.45\textwidth}{!}{
\begin{tabular}{cccc}
\hline
Dataset details & MIRFLICKR-25K & NUS-WIDE & MS COCO \\ \hline
Dataset Size & 24581 & 192779 & 122218 \\
Training Size & 10000 & 10500 & 10000 \\
Retrieval Size & 22581 & 190679 & 117218 \\
Query Size & 2000 & 2100 & 5000 \\
Number of Categories & 24 & 21 & 80 \\ \hline
\end{tabular}}
\par\vspace{-2mm}
\end{table}
\begin{table*}[!htbp]
\centering
\small
\caption{\label{table2}Ablation study results of different components in the proposed method.}
\begin{tabular}{c|c|ccc|ccc|ccc}
\hline
\multirow{2}{*}{} & \multirow{2}{*}{Methods} & \multicolumn{3}{c|}{MIRFLICKR-25K} & \multicolumn{3}{c|}{NUS-WIDE} & \multicolumn{3}{c}{MS COCO} \\ \cline{3-11} 
 &  & 16bit & 32bit & 64bit & 16bit & 32bit & 64bit & 16bit & 32bit & 64bit \\ \hline
\multirow{6}{*}{I2T} & baseline & 0.8444 & 0.8590 & 0.8666 & 0.7228 & 0.7377 & 0.7488 & 0.7099 & 0.7594 & 0.7793 \\
 & w/o (PACL+AGSF) & 0.8767 & 0.8855 & 0.8908 & 0.8278 & 0.8555 & 0.8656 & 0.7222 & 0.7910 & 0.8207 \\
 & w/o (TAAP+PACL) & 0.9139 & 0.9269 & 0.9329 & 0.7668 & 0.7838 & 0.7893 & 0.7117 & 0.7687 & 0.7933 \\
 & w/o AGSF & 0.9346 & 0.9512 & 0.9593 & 0.8377 & 0.8584 & 0.8709 & 0.7424 & 0.8111 & 0.8444 \\
 & w/o PACL & 0.9720 & 0.9859 & 0.9924 & 0.9013 & 0.9450 & 0.9639 & 0.7246 & 0.8228 & 0.8725 \\ \cline{2-11} 
 & PromptHash(Ours) & \textbf{0.9818} & \textbf{0.9960} & \textbf{0.9995} & \textbf{0.9313} & \textbf{0.9759} & \textbf{0.9931} & \textbf{0.7673} & \textbf{0.8782} & \textbf{0.9263} \\ \hline
\multirow{6}{*}{T2I} & baseline & 0.8383 & 0.8488 & 0.8524 & 0.7303 & 0.7477 & 0.7588 & 0.7201 & 0.7595 & 0.7885 \\
 & w/o (PACL+AGSF) & 0.8558 & 0.8641 & 0.8704 & 0.7996 & 0.8272 & 0.8338 & 0.7205 & 0.7804 & 0.8103 \\
 & w/o (TAAP+PACL) & 0.9140 & 0.9276 & 0.9331 & 0.7696 & 0.7841 & 0.7902 & 0.7141 & 0.7755 & 0.7973 \\
 & w/o AGSF & 0.9420 & 0.9591 & 0.9670 & 0.8157 & 0.8313 & 0.8460 & 0.7509 & 0.8092 & 0.8414 \\
 & w/o PACL & 0.9720 & 0.9859 & 0.9925 & 0.9060 & 0.9517 & 0.9676 & 0.7227 & 0.8272 & 0.8731 \\ \cline{2-11} 
 & PromptHash(Ours) & \textbf{0.9816} & \textbf{0.9955} & \textbf{0.9995} & \textbf{0.9381} & \textbf{0.9761} & \textbf{0.9934} & \textbf{0.7667} & \textbf{0.8790} & \textbf{0.9283} \\ \hline
\end{tabular}
\par\vspace{-2mm}
\end{table*}
\begin{table*}[!htbp]
\centering
\caption{\label{table3}For the Mean Average Precision (MAP) evaluation on the MIRFLICKR25K, NUS-WIDE, and MS COCO datasets (MAP@ALL), the best performance is in \textbf{bold}, the second-best is \underline{underlined}, and results from original papers are marked with an asterisk (*).}
\resizebox{\textwidth}{!}{
\begin{tabular}{c|c|c|cccc|cccc|cccc}
\hline
\multirow{2}{*}{} & \multirow{2}{*}{Network} & \multirow{2}{*}{Methods} & \multicolumn{4}{c|}{MIRFLICKR-25K} & \multicolumn{4}{c|}{NUS-WIDE} & \multicolumn{4}{c}{MS COCO} \\ \cline{4-15} 
 &  &  & 16bit & 32bit & 64bit & Avg & 16bit & 32bit & 64bit & Avg & 16bit & 32bit & 64bit & Avg \\ \hline
\multirow{12}{*}{I2T} & \multirow{3}{*}{CNN} & DCMH~\cite{b3} & 0.7288 & 0.7411 & 0.7490 & 0.7396 & 0.5238 & 0.5995 & 0.6195 & 0.5809 & 0.5177 & 0.5311 & 0.5471 & 0.5320 \\
 &  & CMHH~\cite{b11} & 0.6649 & 0.6677 & 0.6731 & 0.6686 & 0.5312 & 0.5476 & 0.5299 & 0.5362 & 0.5657 & 0.5582 & 0.5540 & 0.5593 \\
 &  & GCDH~\cite{b23} & 0.7991 & 0.8123 & 0.8204 & 0.8106 & 0.7142 & 0.7367 & 0.7498 & 0.7336 & 0.7297 & 0.7651 & 0.7905 & 0.7618 \\ \cline{2-15}
 & \multirow{9}{*}{CLIP} & DCHMT~\cite{b24} & 0.8254 & 0.8284 & 0.8257 & 0.8265 & 0.6832 & 0.6892 & 0.7025 & 0.6916 & 0.6440 & 0.6465 & 0.6552 & 0.6486 \\
 &  & MITH~\cite{b25} & 0.8377 & 0.8491 & 0.8576 & 0.8481 & 0.7062 & 0.7180 & 0.7186 & 0.7143 & 0.7149 & 0.7373 & 0.7580 & 0.7367 \\
 &  & DSPH~\cite{b27} & 0.8081 & 0.8387 & 0.8505 & 0.8324 & 0.6953 & 0.7031 & 0.7161 & 0.7048 & 0.7049 & 0.7583 & 0.7728 & 0.7453 \\
 &  & TwDH~\cite{b28} & 0.7766 & 0.7915 & 0.8016 & 0.7899 & 0.6649 & 0.6855 & 0.6933 & 0.6812 & 0.6564 & 0.7207 & 0.7472 & 0.7081 \\
 &  & DNpH~\cite{b29} & 0.8399 & 0.8486 & 0.8500 & 0.8462 & 0.7135 & 0.7169 & 0.7247 & 0.7184 & 0.6754 & 0.6897 & 0.6862 & 0.6838 \\
 &  & DHaPH~\cite{b30} & 0.8271 & 0.8351 & 0.8324 & 0.8315 & 0.7215 & 0.7333 & 0.7410 & 0.7319 & 0.7310 & 0.7402 & 0.7496 & 0.7403 \\
 &  & CMCL~\cite{b31} & 0.8487 & 0.8611 & 0.8692 & 0.8597 & 0.7154 & 0.7314 & 0.7440 & 0.7303 & {\ul 0.7420} & {\ul 0.7701} & {\ul 0.7936} & {\ul 0.7686} \\
 &  & VTPH*~\cite{b9} & {\ul 0.9056} & {\ul 0.9249} & {\ul 0.9328} & {\ul 0.9211} & {\ul 0.7733} & {\ul 0.7870} & {\ul 0.7936} & {\ul 0.7846} & 0.7202 & 0.7477 & 0.7786 & 0.7488 \\
 &  & PromptHash(Ours) & \textbf{0.9818} & \textbf{0.9960} & \textbf{0.9995} & \textbf{0.9924} & \textbf{0.9313} & \textbf{0.9759} & \textbf{0.9931} & \textbf{0.9668} & \textbf{0.7673} & \textbf{0.8782} & \textbf{0.9263} & \textbf{0.8573} \\ \hline
\multirow{12}{*}{T2I} & \multirow{3}{*}{CNN} & DCMH~\cite{b3} & 0.7520 & 0.7696 & 0.7776 & 0.7664 & 0.5440 & 0.5901 & 0.5956 & 0.5766 & 0.5510 & 0.5883 & 0.6050 & 0.5814 \\
 &  & CMHH~\cite{b11} & 0.6615 & 0.6851 & 0.6943 & 0.6803 & 0.4826 & 0.4868 & 0.4711 & 0.4802 & 0.5098 & 0.5071 & 0.5097 & 0.5089 \\
 &  & GCDH~\cite{b23} & 0.7849 & 0.8022 & 0.8067 & 0.7979 & 0.7215 & 0.7423 & 0.7534 & 0.7391 & 0.7261 & 0.7650 & 0.7885 & 0.7599 \\ \cline{2-15}
 & \multirow{9}{*}{CLIP} & DCHMT~\cite{b24} & 0.8115 & 0.8179 & 0.8200 & 0.8165 & 0.6920 & 0.7081 & 0.7208 & 0.7070 & 0.6282 & 0.6361 & 0.7674 & 0.6772 \\
 &  & MITH~\cite{b25} & 0.8288 & 0.8408 & 0.8483 & 0.8393 & 0.7122 & 0.7281 & 0.7335 & 0.7246 & 0.7045 & 0.7319 & 0.7168 & 0.7177 \\
 &  & DSPH~\cite{b27} & 0.7909 & 0.8215 & 0.8334 & 0.8153 & 0.7028 & 0.7165 & 0.7329 & 0.7174 & 0.7047 & 0.7552 & 0.7525 & 0.7375 \\
 &  & TwDH~\cite{b28} & 0.7758 & 0.7895 & 0.8030 & 0.7894 & 0.6719 & 0.7126 & 0.7153 & 0.6999 & 0.6603 & 0.7205 & 0.6990 & 0.6933 \\
 &  & DNpH~\cite{b29} & 0.8151 & 0.8249 & 0.8329 & 0.8243 & 0.7222 & 0.7265 & 0.7313 & 0.7267 & 0.6595 & 0.6801 & 0.6990 & 0.6795 \\
 &  & DHaPH~\cite{b30} & 0.8089 & 0.8170 & 0.8194 & 0.8151 & 0.7203 & 0.7284 & 0.7388 & 0.7292 & 0.6999 & 0.7037 & 0.7919 & 0.7318 \\
 &  & CMCL~\cite{b31} & 0.8290 & 0.8362 & 0.8402 & 0.8351 & 0.7290 & 0.7438 & 0.7511 & 0.7413 & {\ul 0.7396} & {\ul 0.7689} & {\ul 0.7962} & {\ul 0.7682} \\
 &  & VTPH*~\cite{b9} & {\ul 0.9020} & {\ul 0.9226} & {\ul 0.9278} & {\ul 0.9175} & {\ul 0.7708} & {\ul 0.7840} & {\ul 0.7933} & {\ul 0.7827} & 0.7185 & 0.7515 & 0.7743 & 0.7481 \\
 &  & PromptHash(Ours) & \textbf{0.9816} & \textbf{0.9955} & \textbf{0.9995} & \textbf{0.9922} & \textbf{0.9381} & \textbf{0.9761} & \textbf{0.9934} & \textbf{0.9692} & \textbf{0.7667} & \textbf{0.8790} & \textbf{0.9283} & \textbf{0.8580} \\ \hline
\end{tabular}}
\par\vspace{-2mm}
\end{table*}
\subsection{Analysis of Experimental Results}
\subsubsection{Ablation Studies}
The experimental results demonstrate that incorporating the TAAP module significantly enhances retrieval performance by effectively addressing text semantic truncation through adaptive weighting of original text semantic features, thereby preserving retrieval-relevant semantic information while minimizing the impact of irrelevant features. Implementation of the AGSF module reveals the effectiveness of cross-modal semantic fusion in selectively retaining beneficial semantic information while filtering out redundant contextual information. The combination of TAAP and PACL modules demonstrates that PACL's alignment of global and local prompt tokens optimizes prompt text semantics and original text semantics with image semantics as the center, resulting in high-quality hash codes. While results from PACL indicate that TAAP and AGSF modules substantially improve retrieval performance for high-bit hash codes, they are less effective for low-bit scenarios. Notably, the complete PromptHash framework, incorporating all three modules, successfully leverages their complementary strengths to enhance retrieval performance across all hash code lengths, validating the superiority of our designed modules.

\subsubsection{Hyperparameter Analysis}

We conduct a comprehensive analysis of hyperparameters in our proposed PromptHash method, with results illustrated in~\cref{figure3} for the six hyperparameters in our overall objective function. The experimental results demonstrate optimal performance across all datasets when setting $\{\alpha=5.0, \beta=5.0, \gamma=0.005, \mu=5.0, \sigma=0.1, \zeta=0.001\}$. Our analysis reveals that gradually increasing parameters while maintaining default values for others yields optimal mean Average Precision (mAP) across the three datasets. However, reducing $\sigma$ to 0.1 while simultaneously increasing $\gamma$ results in decreased mAP performance. Furthermore, when attempting to optimize all hyperparameters independently for each dataset, the resulting average mAP values prove inferior to those achieved by solely reducing $\sigma$ to 0.1. Additional experiments show that further reducing $\sigma$ to 0.05 also leads to degraded average mAP performance. These findings suggest that maintaining default values for all hyperparameters except the quantization loss weight $\sigma$ (set to 0.1) yields optimal average mAP across all three datasets. We hypothesize that this phenomenon occurs because PromptHash effectively aligns image-text features extracted by CLIP, resulting in high similarity between related image-text pairs and natural dissimilarity between unrelated pairs in the common Hamming space, thus requiring lower emphasis on quantization loss.

\subsubsection{Comparison with State-of-the-Art}

As demonstrated in~\cref{table3}, our proposed PromptHash method achieves superior performance across all metrics on three public datasets. Using the mAP@all metric as a primary indicator, we observe significant improvements over the previous state-of-the-art methods. On the MIRFLICKR-25K dataset, PromptHash outperforms the second-best method (VTPH) by margins of 7.13\% and 7.47\% for I2T and T2I tasks, respectively. For the NUS-WIDE dataset, we achieve even more substantial improvements over VTPH, with gains of 18.22\% for I2T and 18.65\% for T2I tasks. On the MS COCO dataset, our method surpasses the previous best performer (CMCL) by 8.87\% and 8.98\% for I2T and T2I tasks, respectively.

The Precision-Recall curves in~\cref{figure4} demonstrate particularly notable improvements on MIRFLICKR-25K and NUS-WIDE datasets, where PromptHash significantly outperforms existing methods. This superior performance can be attributed to our method's effective handling of discrete word-based texts through affinity prompt learning and adaptive weighting modules, coupled with image-centric feature discrimination between foreground retrieval targets and background information. However, we observe relatively modest improvements on the MS COCO dataset, which we attribute to its unique characteristics: text annotations comprise complete sentences rather than discrete words, with annotations extending up to 625 words, of which only 3-5 words typically represent retrieval targets. This high noise data presents a particular challenge that could potentially be addressed in future work through text reconstruction.
\section{Conclusion}
\label{sec:conclusion}
Most existing cross-modal hashing methods primarily focus on hashing function design and optimizing distances between different samples, while overlooking inherent issues within the sample sets themselves. Specifically, problems such as text semantic truncation and negative semantic information within the sample set can significantly limit retrieval performance. 

Our proposed PromptHash framework addresses these limitations through two key innovations. First, it introduces prompt learning for adaptively weighted learning of truncated text semantics, effectively improving retrieval performance under limited character encoding constraints. Second, it combines SSM and Transformer architectures with adaptive gated selection fusion, enabling efficient filtering and weighting of fused features. Furthermore, our novel Prompt Affinity Contrastive Learning module (PACL) balances prompt, textual, and visual features while bridging modality heterogeneity through feature alignment, significantly enhancing retrieval accuracy.

\section*{Acknowledgments}
This work was supported by the Scientific and Technological Innovation 2030 Major Project under Grant 2022ZD0115800, the National Natural Science Foundation of China under Grant 62441213, and the Key Laboratory Open Projects in Xinjiang Uygur Autonomous Region under Grant 2023D04028.

{
    \small
    \bibliographystyle{ieeenat_fullname}
    \bibliography{main}
}


\end{document}